\documentclass[letterpaper, 10 pt, conference]{ieeeconf}  

\IEEEoverridecommandlockouts                              

\usepackage{amsmath}
\usepackage[table,xcdraw]{xcolor}
\usepackage{amssymb}
\usepackage{graphicx}
\usepackage[hidelinks]{hyperref} 
\usepackage{hhline} 
\usepackage{subcaption} 
\usepackage{siunitx} 
\usepackage{wrapfig} 
\usepackage{floatflt}
\usepackage{booktabs} 

\newcommand{\stddev}[1]{\begin{footnotesize}$\pm$#1\end{footnotesize}}
\newcommand{\mean}[1]{\begin{footnotesize}#1\end{footnotesize}}

\title{\LARGE \bf
SceneScore:\\Learning a Cost Function for Object Arrangement
}

\author{Ivan Kapelyukh$^{1,2}$, Edward Johns$^1$
\thanks{$^{1}$ The Robot Learning Lab at Imperial College London. $^2$ The Dyson Robotics Lab at Imperial College London.
        }%
}

\begin{document}

\maketitle
\thispagestyle{empty}
\pagestyle{empty}
\renewcommand\bottomfraction{.1}
\renewcommand\topfraction{.1}

\begin{abstract}
Arranging objects correctly is a key capability for robots which unlocks a wide range of useful tasks. A prerequisite for creating successful arrangements is the ability to evaluate the desirability of a given arrangement. Our method ``SceneScore'' learns a cost function for arrangements, such that desirable, human-like arrangements have a low cost. We learn the distribution of training arrangements offline using an energy-based model, solely from example images without requiring environment interaction or human supervision. Our model is represented by a graph neural network which learns object-object relations, using graphs constructed from images. Experiments demonstrate that the learned cost function can be used to predict poses for missing objects, generalise to novel objects using semantic features, and can be composed with other cost functions to satisfy constraints at inference time. Videos are available at: \href{https://sites.google.com/view/scenescore}{sites.google.com/view/scenescore}.
\end{abstract}
\section{Introduction}
\label{sec:intro}

\begin{figure}[ht]
    \centerline{\includegraphics[width=0.7\linewidth]{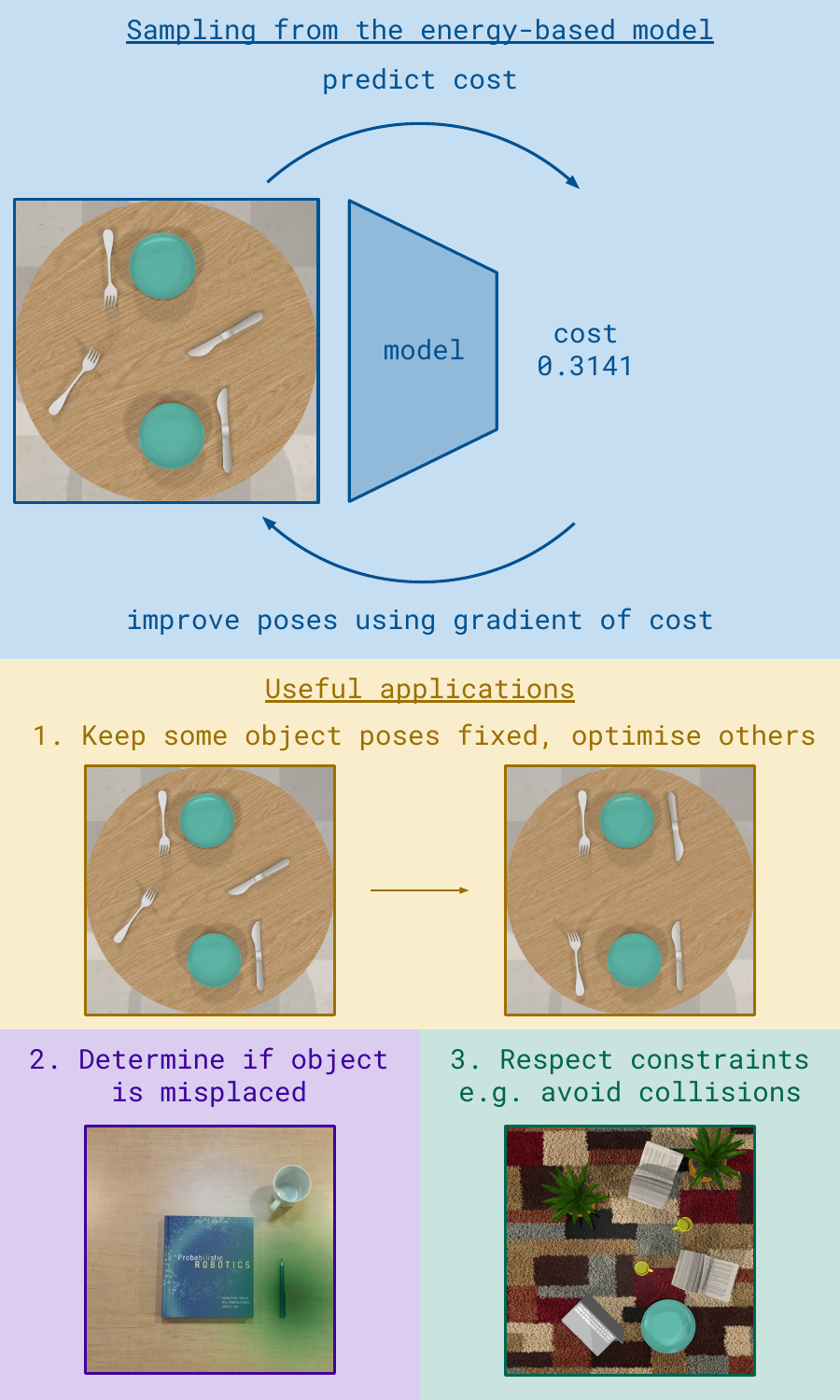}}
    \caption{An overview of how to sample from our energy-based model and several useful properties of implicit methods.}
    \label{fig:train-infer}
\end{figure}

Object rearrangement is a ubiquitous challenge in robotics: given a set of objects, arrange them into a desirable state \cite{rearrangement}. Many tasks can be expressed as rearrangement problems, e.g. tidying a room, loading a dishwasher, assembling furniture in a factory, or setting a table in a restaurant. For a robot to be proficient at rearrangement in the real world, it must be able to evaluate the desirability of an arrangement. Our method ``SceneScore'' learns a cost function for arrangements, such that desirable, human-like arrangements have a low cost, and random arrangements have a higher cost. This cost function can then be minimised to determine low-cost target poses for objects (Fig. \ref{fig:train-infer}). One application of this method is to provide target poses for robotic systems which can physically perform the rearrangement \cite{ifor,wheretostart,cabinet}.

Prior work \cite{neatnet,structformer} has explored the problem of predicting an optimal arrangement for a set of objects. This is a related but different problem. By learning a cost function, our method can also be used to find an optimal arrangement. However, learning a cost function for any given arrangement (an \textit{implicit} approach) gives our method several advantages compared to methods which predict only the optimal arrangement (an \textit{explicit} approach). These benefits include:\\
\textbf{(1) Compositionality}. The learned cost function can be composed with additional cost functions at inference time. E.g. the time it would take the robot to create an arrangement can be added to the desirability cost of that arrangement. The composed cost function can then be differentiated to efficiently find an arrangement which is both desirable and can be created quickly. Explicit methods cannot be easily composed without re-training, as discussed in \cite{ebms-for-imgs}.
\newpage

\textbf{(2) Robustness}. This is a consequence of compositionality. In real-world scenes, there are many physical constraints which must be satisfied involving robot joints, object collisions, stacking stability, etc. Explicit methods are less robust, since the predicted optimal arrangement may violate constraints. The cost function approach enables the use of efficient gradient-based constrained optimisation to determine a high-quality arrangement which satisfies these constraints. \\
\textbf{(3) Theoretical advantages}. Prior work \cite{ibc} has shown that implicit methods have theoretical advantages over explicit methods, such as better handling of discontinuities and stronger generalisation.

When humans evaluate the quality of an arrangement, many factors are considered, such as aesthetics, convenience, and stability. We investigate learning this function from images of human-arranged scenes. This is a scalable research direction as images of tidy desks and loaded dishwashers are abundant on the Web. Unsupervised learning from this web-scale data could lead to generalist cost functions for object arrangement, analogous to successes in other fields \cite{imagen,gpt3}.
    
We learn the distribution of example arrangements using an energy-based model, which can successfully learn complex distributions \cite{ebms-for-imgs,ibc}. However, learning the distribution from images directly is difficult due to their high dimensionality. Instead, we first create an object-centric graph representation of scenes, which separates an object's pose and semantic features. A key insight is that this allows us to make the cost function conditional: the semantic identities of the objects must remain fixed, but the robot is free to vary the poses in order to create a higher-quality arrangement.

The main contributions of this paper are as follows:

\begin{itemize}
    \item
    \textbf{An algorithm for learning the distribution} of example arrangements using techniques from energy-based modelling, which enables a cost function to be learned solely from example object arrangements, without also requiring environment interaction or human supervision.

    \item
    \textbf{A graph neural network architecture} for predicting the cost of an arrangement, with relative poses as edge features and pose-invariant semantic embeddings as node features. This object-centric, abstracted representation which separates pose and semantics makes generating target arrangements offline possible.

    \item
    \textbf{A vision pipeline for creating graphs} from images of scenes, using a pre-trained web-scale \textbf{CLIP} \cite{clip} model to obtain visual and semantic features for each object, thus allowing generalisation to new objects.
\end{itemize}

To the best of our knowledge, this is the first method which learns a rearrangement cost function from images of scenes, without any annotations. Please visit our website for videos, code, and supplementary materials with real-world demos: \href{https://sites.google.com/view/scenescore}{sites.google.com/view/scenescore}.


\section{Related Work}

We now summarise several common approaches to learning desirable arrangements, followed by a brief overview of energy-based models, the basis of our approach.

\textbf{Classification methods} select target poses from a fixed set of discrete choices. Generalisation to novel objects can be achieved using object taxonomies \cite{taxonomies} or language models \cite{housekeep,tidybot}. A graph neural network \cite{gnn-policy,tidee} or transformer \cite{adaptable-planners} policy can select the correct goal for each object. This can be trained with reinforcement learning to complete long-horizon rearrangement tasks \cite{block-assembly}. While these are effective for high-level task planning, our method predicts a precise \textit{continuous} target position and orientation for each object. Some methods do predict dense pixel-wise location scores \cite{transporters,cliport}, but require full rearrangement demonstrations.

\textbf{Explicit methods} directly predict target poses for each object. These solve a related but different problem to learning a cost function. NeatNet \cite{neatnet} models user preferences as latent vectors by training as a Variational Autoencoder on scenes. However, this model only uses object names as semantic embeddings, whereas our method also uses visual features. Point clouds can also be used to represent objects as shown in \cite{structformer}, where an autoregressive model conditioned on language commands is used to place objects into the scene. Another line of work takes a ``denoising'' approach to determine goal arrangements \cite{structdiffusion,targf,legonet}. While this helps avoid collisions in rearrangement, it addresses a different problem: we are interested in developing an implicit method which, given an arrangement, can \textit{evaluate} how desirable it is.

\textbf{Zero-shot approaches} can use large language (or visual-language) models trained on web-scale data to predict desirable target poses without training on example arrangements \cite{dall-e-bot,dream2real,llm-rearrange}. Instead, SceneScore allows users to provide their own examples which reflect their needs.

\textbf{Energy-based models} (EBMs) are generative models which approximate the training distribution using an unnormalised energy function. Intuitively, the energy function is usually ``pushed down'' at training examples, and ``pushed up'' elsewhere. They have been used for generating high-dimensional images \cite{ebm-high-res}. They have also been used for learning visual and spatial concepts \cite{ebm-concept-learning}, and for learning from demonstrations \cite{ibc,implicit-action-space}. Recent work on EBMs for rearrangement \cite{ebm-relations} demonstrates compositionality by training a separate EBM for every spatial relation (such as \textit{``left of''} or \textit{``line''}), and composing them together to comply with detailed user instructions. Our work investigates a different problem: learning to evaluate autonomously whether an arrangement of a scene looks natural, without requiring detailed user instructions. This leads to several differences in our approaches. For example, in Section \ref{ssec:exp-missing}, we train a single EBM to learn the joint distribution of object poses (including orientations), capturing what it means for a dining table to be conveniently set for human use. This avoids the need to train separate EBMs for each low-level spatial relation, or to collect datasets of scenes annotated with these spatial relations. Additionally, our EBM is conditioned on per-object visual CLIP features (rather than just object positions and sizes), allowing it to take object semantics into account. We also show how the performance advantage of implicit methods grows over explicit methods as collision constraints become more restrictive.

\section{Method}

First we formulate learning a rearrangement cost function as a density estimation problem (Section \ref{ssec:formulating-problem}). Then, we describe a method for learning this density function from example arrangements (Section \ref{ssec:training-ebm}), and show how to sample arrangements from this model (Section \ref{ssec:sample-ebm}). Next, we detail the graph neural network architecture acting as our model (Section \ref{ssec:gnn-arch}). Finally, we illustrate how these graphs can be constructed from images of scenes (Section \ref{ssec:construct-graphs}).

\subsection{Formulating the Problem}\label{ssec:formulating-problem}

To approximate the function that humans use to assess the quality of a scene, we use a neural network model $E_\theta$, with learned parameters $\theta$. The architecture is described in Section \ref{ssec:gnn-arch}. We use the notation $E_\theta$ because this network is trained as an EBM. The input to $E_\theta$ is a representation of a scene, and the output is a scalar cost.

A scene is represented as an object-centric graph (the input to our model). Each object in a scene is represented by its pose and its semantic embedding. The semantic embedding is a vector which captures visual and semantic features of an object that are useful for arranging it, for example its shape, or its semantic function (e.g. cutlery is often placed together in a drawer). This graph is constructed from an image, as described in Section \ref{ssec:construct-graphs}. Let $x$ denote the set of absolute pose vectors for all the objects in a scene, and $s$ the set of semantic embeddings. Then the cost of the scene according to our model is $E_\theta(x|s)$. This separation between pose and semantics is crucial under the definition of rearrangement: the robot is allowed to vary the pose of objects $x$, but it cannot alter or discard the objects themselves, so $s$ must remain fixed.

\subsection{Training the Energy-Based Model}\label{ssec:training-ebm}

We fit the model to the training examples using Maximum Likelihood Estimation (MLE). The probability of an arrangement under the distribution learned by our model is defined as:
\begin{align}\label{eq:ebm-pdf}
    p_\theta(x|s) &= \frac{e^{-E_\theta(x|s)}}{Z_\theta}
    & Z_\theta &= \int_x e^{-E_\theta(x|s)} \text{d}x
\end{align}

 This probability definition is widely used in EBM literature \cite{ebms-for-imgs,ibc,train-ebms,ebm-irl-connection}. Arrangements with a lower cost have a higher probability. The MLE loss is obtained by minimising the negative log-likelihood. We cannot compute an integral over all possible arrangements, but we can approximate the normalisation effect of $Z_\theta$ by sampling arrangements from our learned distribution $p_\theta$. The sampling algorithm is described in Section \ref{ssec:sample-ebm}. Replacing $Z_\theta$ using the training example $i$ and the sample arrangements indexed by $j$, we get:

\begin{align}\label{eq:mle-loss}
    L_\text{MLE}(\theta) = \sum_i -\log\left(\frac{e^{-E_\theta(x_i|s_i)}}{e^{-E_\theta(x_i|s_i)} + \sum_j e^{-E_\theta(x_j|s_i)}}\right)
\end{align}

This takes the form of an InfoNCE-style loss \cite{info-nce}, analogous to those used for EBM training in \cite{ibc,implicit-action-space}. Note that $s_i$ = $s_j$, i.e. the samples $j$ all have the same semantic embeddings as the training example $i$. Intuitively, this loss function encourages the model to assign a low cost (i.e. high probability) to the training examples, and a high cost (i.e. low probability) to the generated arrangements which have been sampled from the model -- they can be viewed as counter-examples. This creates a high-probability region around the training examples, and ``pushes down'' the probability elsewhere.

\subsection{Sampling from the Energy-Based Model}\label{ssec:sample-ebm}

We need to sample from the learned distribution $E_\theta(x|s)$ to approximate the normalising constant in the loss function in (\ref{eq:mle-loss}). To sample poses for a given set of objects with semantic embeddings $s$, we use Langevin Dynamics \cite{langevin}, which is often used to sample from energy-based models \cite{ebms-for-imgs,ibc}. The initial poses $x_0$ are drawn randomly from a uniform distribution. The poses are then updated in each step $t$:
\begin{align}\label{eq:langevin}
    x_t &= x_{t-1} - \lambda_t (\nabla_{x_{t-1}} E_\theta(x_{t-1} | s) + \omega_t) & \omega_t &\sim \mathcal{N}(0, \sigma_t^2)
\end{align}

We take the final poses $x$ as the sampled arrangement, following \cite{langevin}. Here, $\nabla_{x_{t-1}}$ means taking the gradient of the cost function with respect to the object poses. $\lambda_t$ is the step size. This update rule is similar to gradient descent, but with an added noise term $\omega_t$. Langevin Dynamics is run for a finite, fixed number of steps. Further details about hyperparameters are in the supplementary material. At inference time, we also use Langevin Dynamics to sample low-cost arrangements.

\begin{figure*}[htb]
    \centerline{\includegraphics[width=0.8\textwidth]{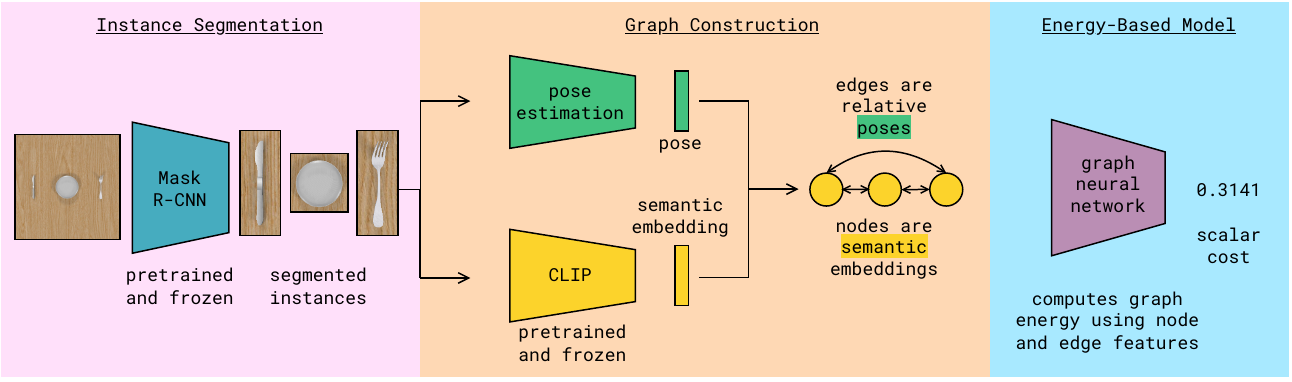}}
    \caption{The pipeline for computing a cost from an image of a scene.}
    \label{fig:vision-pipeline}
\end{figure*}

\subsection{Graph Neural Network Architecture}\label{ssec:gnn-arch}

$E_\theta(x|s)$ is represented by a graph neural network (GNN). Each scene is a fully-connected graph with a node for each object. The edge between two object nodes represents the relative displacement and orientation between them. GNNs are well-suited for this task because they can handle inputs with a variable number of nodes, and they can capture complex, multi-modal training distributions, e.g. if there are multiple acceptable poses for an object. An insight we make is that relative poses between objects often matter more than their absolute pose in the scene, e.g. a pair of slippers being placed together. This motivates our use of relative poses as edge features in the graph, compared to prior work \cite{neatnet} which uses absolute coordinates in the node feature vectors. When sampling, we convert the absolute poses $x$ to relative poses, use the GNN to compute the cost of the scene $E_\theta(x|s)$, and then improve the absolute poses using (\ref{eq:langevin}).

We now describe how one layer of our GNN computes the output features for each node, which are used as input features to the next layer. The input features in the first layer are the semantic embeddings. The edge features stay the same at each layer of the network. For node $i$, the input feature vector is $v_i$, and the output feature vector is $v_i'$:
\begin{align}
    v_i' = \sum_j f_\phi \left( v_i, v_j, e_{ji} \right)
\end{align}
For each neighbouring node $j$ in the graph, we compute a message from $j$ to $i$, and then aggregate all these messages to produce the output feature vector for node $i$. To compute this message, the feature vectors of nodes $i$ and $j$ are concatenated together, along with the edge feature vector $e_{ji}$, which represents the pose transformation to get from the pose of $j$ to the pose of $i$. This concatenated vector is then passed through a linear neural network layer with learned parameters $\phi$. The GNN consists of several of these graph layers, each separated by a LeakyReLU non-linearity \cite{leaky-relu}. We use global add pooling to aggregate the node feature vectors into a single graph encoding vector. This is passed through several linear layers, the output of which is a scalar, which is the cost $E_\theta(x|s)$.

\subsection{Constructing Graphs from Images}\label{ssec:construct-graphs}

The GNN takes as input a graph representing a scene. The system for constructing this graph from an image, and obtaining a cost for this graph, is shown in Fig. \ref{fig:vision-pipeline}. First, our method detects objects in the image. We use a pre-trained Mask R-CNN \cite{mask-rcnn} from the detectron2 library \cite{detectron2}. For each object instance, it returns a segmentation mask. We found that our method works even for objects not in the Mask R-CNN training dataset, as long as the mask has an approximately correct shape.

Next, the pose for each object should be estimated. Our method makes it easy to use any existing pose estimation component, including 6-DoF pose estimators. In our experiments, we focus on tabletop scenes where the methods should predict the $x$ and $y$ position of each object, along with a single angle $\theta$ along the axis perpendicular to the tabletop. We use a straightforward method for pose estimation based on image moments, which also applies to novel objects. Further details are in the supplementary material. The pose vector for an object is $(x, y, \cos \theta, \sin \theta)$.

To derive an object's semantic embedding we use features from a pre-trained CLIP model \cite{clip}. This takes as input an image of the object and returns a 512-dimensional CLIP vector which captures visual features, as well as semantics: e.g. a fork and chopsticks may share some semantic features, which is useful as they are arranged in a related way. The semantic embedding of an object should be pose-invariant: the pose is separate so that we can optimise it. Therefore, we rectify and crop the image of the object before inputting it to CLIP. Details are in the supplementary material. Although the CLIP model weights are frozen, we train a semantic embedding extractor end-to-end, which is a 2-layer MLP that extracts useful features from the CLIP vector. However, when we pre-process an object image to be used as CLIP input, we lose information about the scale of that object, which may be useful for arranging it (e.g. ordering by size). To preserve this information, we compute a scale descriptor (the width and height of the object in its rectified pose), and append this to the output of the semantic embedding extractor to create the object's semantic embedding.


\section{Experiments}

We investigate the following research questions: How accurately can our method predict poses for missing objects (Section \ref{ssec:exp-missing})? Can our method generalise to novel objects (Section \ref{ssec:exp-novel})? Can the learned cost function be composed with constraints at inference time (Section \ref{ssec:exp-composing})? Further qualitative results are available on our website: \href{https://sites.google.com/view/scenescore}{sites.google.com/view/scenescore}.

\subsection{Experimental Setup}\label{sec:exp-setup}

We create a dataset of images of arranged scenes in simulation, in order to compare predicted poses against ground-truth poses. We use a data-generating process with ground truth distributions, e.g. sampling relative poses between objects from a Gaussian Mixture Model. Details are in the supplementary materials, along with the full datasets. The methods must infer this distribution from example images alone, which are created by rendering the example scenes in a simulator.

\subsection{Predicting Poses for Missing Objects}\label{ssec:exp-missing}

In this experiment, the method is shown an arranged scene at inference time, with an object missing. The method must predict the correct pose for the missing object, taking into account the poses of the pre-placed objects, and using its learned object-object relations. We compare the following methods:

\begin{figure}[htb]
    \centerline{\includegraphics[width=1.0\linewidth]{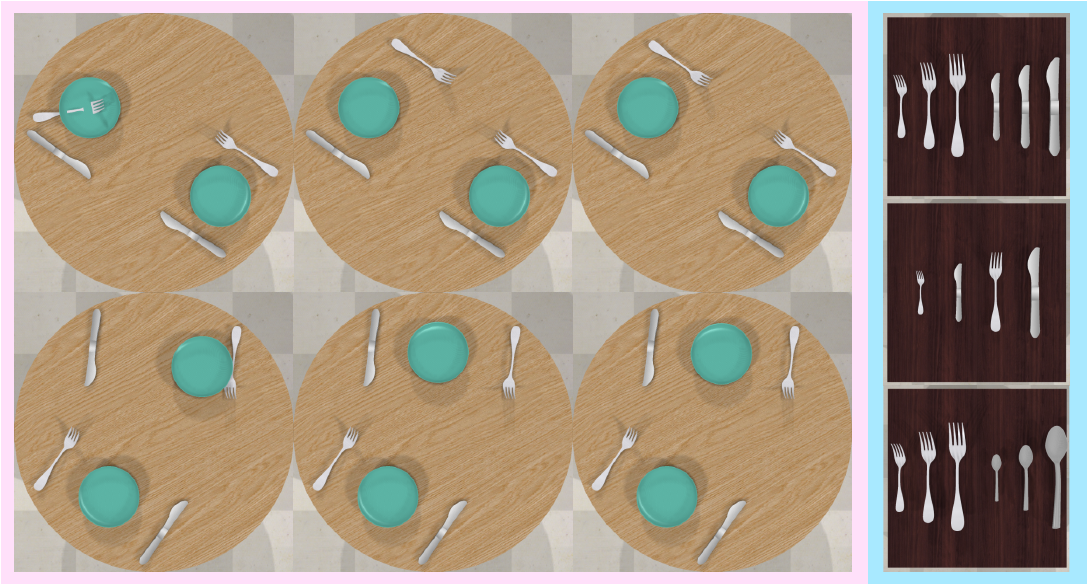}}
    \caption{\textbf{Left:} results for placing missing objects. Top: placing fork, bottom: placing bowl. From left to right, methods are: \texttt{Nearest-Nbr}, \texttt{SceneScore}, ground truth. \textbf{Right:} experiment which requires ordering novel objects.}
    \label{fig:missing-obj}
\end{figure}

(1) \texttt{SceneScore}. Our method learns the distribution of training arrangements. At inference time, the poses of the pre-placed objects are fixed. The missing object's position is randomly initialised, and then optimised using Langevin Dynamics based on the learned cost function. (2) \texttt{SceneScore-Abs}. This ablation study is similar to our method, except that absolute poses are included as node features instead of relative poses in edge features. (3) \texttt{Nearest-Nbr}. This baseline compares the poses of pre-placed objects against each training arrangement to find the closest match, and from that training arrangement returns the pose of the object which is missing from the test arrangement. We expect it to perform well in this experiment because the set of objects is fixed, however it cannot generalise to novel objects. (4) \texttt{NeatNet} \cite{neatnet}. We compare against this prior work for arranging objects, which trains a masked VAE. At inference time, the pose of the missing object is masked out, and is predicted by the decoder. For a fair comparison, we first extend this method to handle orientation. Further implementation details are in the supplementary materials.

For this experiment, we focus on a dining table scenario (Fig. \ref{fig:missing-obj} (left)). There are a pair of plates which are placed circularly at any angle around a table. Beside each plate, a fork and knife are placed an approximately equal distance apart, where this distance varies. To test the methods on multi-modal distributions, the fork and knife can be placed on either side for each arrangement. Together, this creates a challenging task because the methods have to learn relative poses in a circular arrangement, handle multi-object relations, place multiple instances of the same class, and learn multi-modal distributions. There are 48 arrangements for training and 16 for testing.

As shown in Table \ref{tab:missing}, our method \texttt{SceneScore} out-performs the baselines, meaning that the cost function is learned correctly. Using relative poses as edge features improves performance, as shown by the comparison with the \texttt{SceneScore-Abs} ablation. \texttt{Nearest-Nbr} is not able to generalise well to novel configurations at inference time. \texttt{NeatNet} struggles to learn relative poses in the complex circular arrangements, and to tell apart instances of the same class, showing that \texttt{SceneScore}'s EBM approach is better able to learn these distributions and generalise to new configurations.

\subsection{Generalising to Novel Objects}\label{ssec:exp-novel}

We now test the method's ability to use semantic features to generalise to novel objects. A common rearrangement task is to categorise and order objects: laying out cutlery in the kitchen, stacking plates in order of size, etc. We use this setting in this experiment (Fig. \ref{fig:missing-obj} (right)). The methods are shown training examples where objects of varying size and class are ordered according to some rule, which must be learned by the method from the training images. At test time, the method must jointly optimise the poses of a set of novel objects by generalising the learned rule to them. In the first scenario, the objects are first categorised according to class (fork vs knife), and then ordered according to size within each class. In the second scenario, the objects are all ordered according to size, regardless of class. At test time, the forks and knives are of different sizes to those seen during training. In the third scenario, the model only sees knives and forks during training, arranged as in the first scenario. At inference time, it must generalize to an \textit{unseen class}, i.e. spoons, which appear alongside forks in the test scene. This is challenging because it must group the unseen objects together and generalize the learned ordering pattern to them. There are 16 training and 16 test scenes for each scenario. We compare \texttt{SceneScore} against \texttt{NeatNet-R} \cite{neatnet}, which learns to predict poses from semantic embeddings via regression, where the word embedding for each object is derived from the Mask R-CNN class output.

The results are in Table \ref{tab:novel}. Our method \texttt{SceneScore} is better able to generalise to novel objects. \texttt{NeatNet-R} can categorise objects by class using language, but our method allows for more precise placement because it also includes visual features in its semantic embeddings. The Unseen-Class task is more challenging for both methods, causing higher-variance results. Our method uses CLIP features and can generalise the learned object relations to scenes with an unseen class.

\begin{table*}[h]
    \centering
    \sisetup{uncertainty-mode=separate}
    \begin{tabular}{lccccccc}
        \toprule
        Method & Bowl & \multicolumn{2}{c}{Fork} & \multicolumn{2}{c}{Knife} & \multicolumn{2}{c}{Mean} \\
         & $t$ & $t$ & $R$ & $t$ & $R$ & $t$ & $R$ \\
        \midrule
        {\scriptsize \texttt{NeatNet} \cite{neatnet}} & \mean{17.6}\stddev{0.7} & \mean{24.1}\stddev{1.8} & \mean{147.8}\stddev{21.5}  & \mean{23.6}\stddev{1.4} & \mean{145.4}\stddev{23.5} & \mean{21.7} & \mean{146.6} \\
        {\scriptsize \texttt{Nearest-Nbr}} & \mean{5.4}\stddev{3.4} & \mean{9.9}\stddev{9.0} & \mean{16.0}\stddev{11.8} & \mean{8.4}\stddev{8.0} & \mean{17.9}\stddev{11.5} & \mean{7.9} & \mean{16.9} \\
        {\scriptsize \texttt{SceneScore-Abs}} & \mean{5.2}\stddev{6.6} & \mean{6.3}\stddev{5.3} & \mean{9.1}\stddev{4.5} & \mean{7.8}\stddev{8.7} & \mean{20.4}\stddev{40.0} & \mean{6.5} & \mean{14.8} \\
        {\scriptsize \texttt{SceneScore}} & \textbf{\mean{3.4}}\stddev{2.3} & \textbf{\mean{4.1}}\stddev{2.4} & \textbf{\mean{6.4}}\stddev{4.9} & \textbf{\mean{5.6}}\stddev{4.3} & \textbf{\mean{13.0}}\stddev{17.9} \ & \textbf{\mean{4.4}} & \textbf{\mean{9.7}} \\
        \bottomrule
    \end{tabular}
    \caption{Placing missing objects. For each object, the distance error $t$ in cm between the predicted and true positions is shown, followed by the orientation error $R$ in degrees (excluding the bowl due to rotational symmetry).}
    \label{tab:missing}
    \vspace{-0.3cm}
\end{table*}

\begin{table}[htb]
    \centering
    \begin{tabular}{ccccc}
        \toprule
        Method & Class-Size & All-Size & Unseen-Class & Mean \\
        \midrule
        {\scriptsize \texttt{NeatNet-R}\cite{neatnet}} & \mean{6.55}\stddev{4.35} & \mean{12.95}\stddev{6.84} & \mean{17.30}\stddev{11.24} & \mean{12.27} \\
        {\scriptsize \texttt{SceneScore}} & \textbf{\mean{2.89}}\stddev{1.53} & \textbf{\mean{2.01}}\stddev{0.94} & \textbf{\mean{4.76}}\stddev{3.07} & \textbf{\mean{3.22}} \\
        \bottomrule
    \end{tabular}
    \caption{Mean and standard deviation of position error in cm for ordering novel objects.}
    \label{tab:novel}
    \vspace{-0.2cm}
\end{table}

\subsection{Composing the Learned Cost Function with Constraints}\label{ssec:exp-composing}
In this experiment we investigate how constraints can be incorporated into a rearrangement method. This is particularly important for realistic, cluttered scenes. We compare two approaches: an implicit method (ours) which samples a solution and then performs gradient-based constrained optimisation, and an explicit method (NeatNet \cite{neatnet}) which samples solutions and rejects those that violate constraints. We focus on object-object collision avoidance constraints, but this approach can also be used with constraints such as how far the robot arm can reach along a table.

\begin{figure}[htb]
    \centerline{\includegraphics[width=1.0\linewidth]{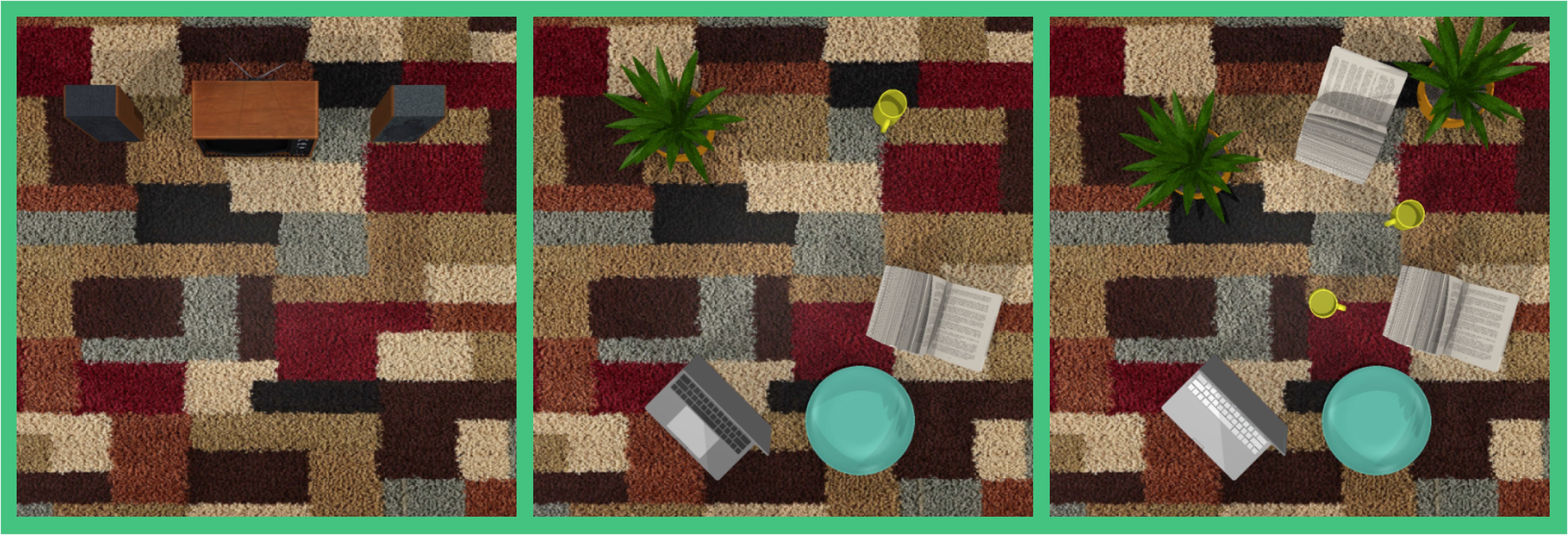}}
    \caption{\textbf{Left:} a training example for arranging the television and two stereo speakers. \textbf{Centre:} clutter added at inference time. \textbf{Right:} even more challenging clutter.}
    \label{fig:tvs}
    \vspace{-0.2cm}
\end{figure}

Suppose that the robot learns a model which evaluates the quality of a television and stereo speaker setup on a living room floor, as shown in Fig. \ref{fig:tvs}. For a balanced audio experience, all three objects should be aligned vertically in a straight line. Additionally, the setup should be horizontally symmetrical, such that the left and right speakers are equidistant from the television. During training, the methods are provided with 36 training arrangements, generated using a similar process to Section \ref{sec:exp-setup}. They must learn that distribution of high-quality arrangements. At inference time, the methods must sample quality arrangements which also satisfy constraints that prevent any objects from colliding. Furthermore, there are now several new objects placed in the scene as clutter, which are included in the constraints, but cannot be moved by the method. These clutter objects are excluded from the graph before pose prediction, since they cannot be moved and were not trained on. Each method is allocated an equal budget of 5000 samples they are allowed to draw, and we report the number of correct samples generated by each method. A correct sample is defined as one which has alignment of objects within a fixed threshold, and contains no object collisions. Collisions are detected from the segmentation masks for objects, but in future work a model such as CollisionNet \cite{collisionnet} can be used. Further details are in the supplementary material.

For the \texttt{NeatNet} \cite{neatnet} baseline, we sample arrangements from the latent space of the VAE and reject the samples that violate the constraints.

\begin{figure}[htb]
    \centerline{\includegraphics[width=0.8\linewidth]{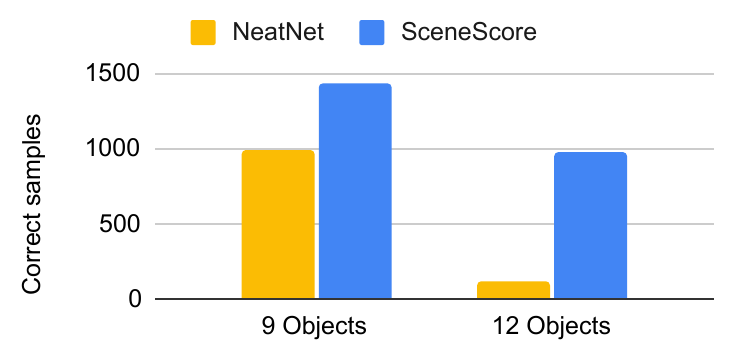}}
    \caption{Number of correct samples from each method as the total number of objects in the scene increases and collisions become harder to avoid.}
    \label{fig:tv-chart}
\end{figure}

For the \texttt{SceneScore} method, the samples are drawn using Langevin Dynamics, as before: the arrangements are initialised randomly and then optimised using the gradient of the learned cost function, summed with the gradient of the constraint function. To make the constraint function differentiable, we use a standard Hinge Loss, which is zero when two objects are not in collision, and linearly increases as the object overlap increases if there is a collision.

The results are in Fig. \ref{fig:tv-chart}. \textbf{As the scene becomes more cluttered and complex, the performance advantage of our implicit method increases over explicit methods} such as NeatNet. Although it is possible for a solution to be sampled from the VAE which luckily satisfies all the constraints, our gradient-based constrained optimisation approach scales much better as the constraints become more challenging, as is the case in complex real-world scenes.


\section{Real-World Demo}

\begin{figure}[htb]
    \centerline{\includegraphics[width=0.7\linewidth]{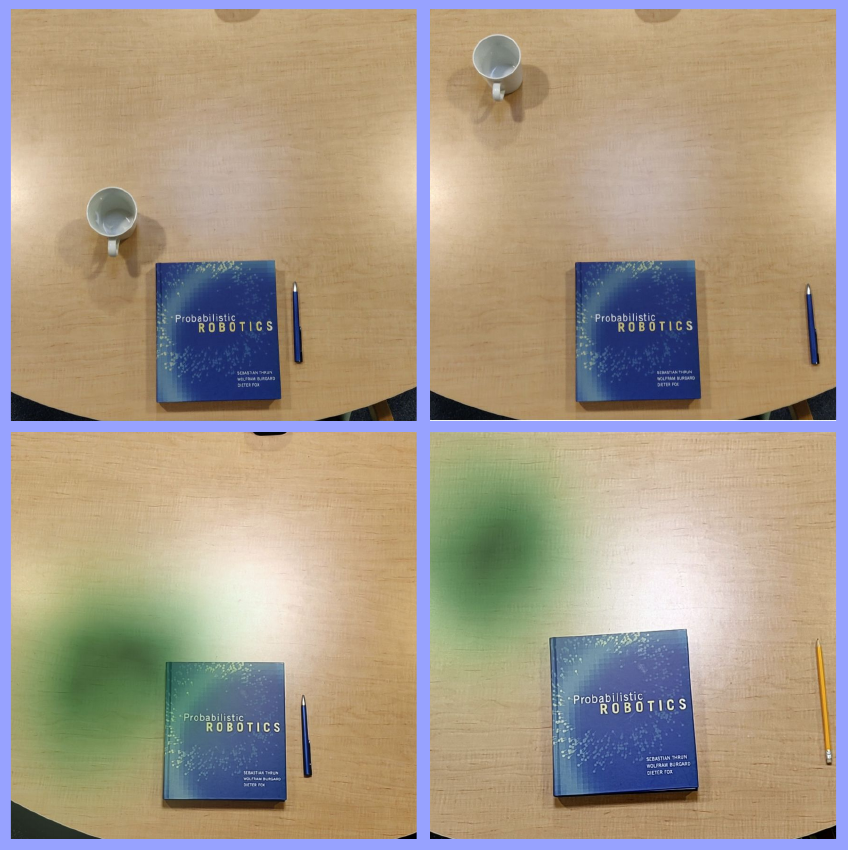}}
    \caption{Visualising the learned cost function for placing a mug in a desirable way, by conditioning the joint distribution on the poses of the other objects.}
    \label{fig:real}
\end{figure}

We conduct a small-scale demo of our method on a real scene, and visualise qualitative results in Fig. \ref{fig:real}. As the pen moves further from the book in the training examples, the mug moves further as well. By fixing the poses of the other objects (book or pen) and visualising the cost at each mug position, we can see that the model has learned this spatial relation. Additionally, it can generalise this relation when the pen is replaced with an unseen class (a pencil), using semantic features. This also shows how to obtain a conditional distribution from the learned joint distribution. Further details are in the supplementary material.


\section{Conclusions}

\textbf{Findings}. Our method SceneScore learns the distribution of example arrangements from images, using an EBM and a graph representation of scenes. The learned cost function can be used to create low-cost arrangements, generalise to novel objects, and can be composed with constraints at inference time. This is the first method which learns a cost function for arranging objects from unannotated images.\\
\textbf{Limitations \& future work}. We currently focus on top-down arrangements. This is sufficient to solve many rearrangement tasks, but future work can also apply our method in a 3D context, e.g. with shelves. Our experiments are mostly in simulation, since we want to compare our predictions against ground truth poses. There are small-scale real-world demos on our website, but future work can investigate this approach in more complex real-world scenes. A promising direction for future work based on this approach is web-scale learning from in-the-wild images of scenes arranged by humans.


\clearpage
\addtolength{\textheight}{-12cm}   




\section*{Acknowledgements}

We thank Andrew Davison, Ignacio Alzugaray, Tristan Laidlow, Edgar Sucar, and Vitalis Vosylius for helpful discussions. This work was supported by Dyson Technology Ltd, and the Royal Academy of Engineering under the Research Fellowship Scheme.
\vspace{-0.2cm}


\bibliographystyle{IEEEtran}
\bibliography{IEEEabrv, scenescore}

\end{document}